\documentclass{bmvc2k}

\title{FurnSet: Exploiting Repeats for 3D Scene Reconstruction}

\addauthor{Paul Dobre}{paul.dobre@ucalgary.ca}{1}
\addauthor{Xin Wang}{xcwang@ucalgary.ca}{1}
\addauthor{Hongzhou Yang}{honyang@ucalgary.ca}{2}

\addinstitution{
 University of Calgary\\
 Calgary, Alberta, Canada
}

\runninghead{DOBRE ET AL. }{FURNSET}

\usepackage{float}
\usepackage{booktabs}
\usepackage{multirow}
\usepackage{amsmath}
\usepackage{amssymb}

\begin{document}

\maketitle

\begin{abstract}
Single-view 3D scene reconstruction involves inferring both object geometry and spatial layout. Existing methods typically reconstruct objects independently or rely on implicit scene context, failing to exploit the repeated instances commonly present in real-world scenes. We propose FurnSet, a framework that explicitly identifies and leverages repeated object instances to improve reconstruction. Our method introduces per-object CLS tokens and a set-aware self-attention mechanism that groups identical instances and aggregates complementary observations across them, enabling joint reconstruction. We further combine scene-level and object-level conditioning to guide object reconstruction, followed by layout optimization using object point clouds with 3D and 2D projection losses for scene alignment. Experiments on 3D-Future and 3D-Front demonstrate improved scene reconstruction quality, highlighting the effectiveness of exploiting repetition for robust 3D scene reconstruction.
\end{abstract}

\section{Introduction}
\label{sec:intro}
3D scene reconstruction from single-view images has received significant attention due to its vast applicability across domains. With the rapid development of augmented reality (AR), extended reality (XR), and simulation for robotics and autonomous driving, there is an increased demand for recovering accurate and high-fidelity 3D scene structure from images. Despite this demand, high-quality 3D data remains scarce relative to the abundance and diversity of 2D images \cite{Tang2025Survey,Zhang2025SurveyFF}. Reconstructing a 3D scene from a single image involves two key challenges: inferring the geometry of individual objects and estimating their spatial layout within the scene. Recent methods for single-view 3D scene reconstruction typically fall into two categories for recovering object geometry: retrieving 3D objects from a CAD database \cite{Wu2025Diorama, Gao2024DiffCAD,Siddiqui2021RetrievalFuse,Gumeli2022ROCA,Avetisyan2019Scan2CAD} or generating object geometry conditioned on the input image \cite{Meng2025SceneGen,Huang2025MIDI,Yao2025CAST,Zhao2025DEPR}. While retrieval-based methods deliver high-quality meshes that can be directly used in downstream applications, their generalization and diversity are inherently limited by the CAD database. Generative approaches instead synthesize object geometry directly from the input image, enabling models to capture the diversity of real-world objects without relying on a fixed database of shapes \cite{Huang2025MIDI,Meng2025SceneGen,Yao2025CAST,Zhao2025DEPR}. After reconstructing object geometry, the spatial placement of objects within the scene must also be estimated. Some approaches jointly predict object geometry and spatial placement in a single model \cite{Huang2025MIDI,Meng2025SceneGen, Chen2025SAM3D} while others decompose the problem into separate modules for geometry reconstruction and layout estimation \cite{Zhai2023CommonScenes, Nie2020Total3DUnderstanding,Yao2025CAST,Zhao2025DEPR}. Additionally, works further leverage vision-language models (VLMs) to complement layout estimation by incorporating semantic understanding of objects and their relationships within the scene \cite{Wu2025Diorama,Yao2025CAST,Sun2025LayoutVLM}.

A significant challenge in 3D scene reconstruction is the inherently amodal nature of the task, as objects in the scene frequently occlude one another. Consequently, both object geometry and spatial placement must be inferred despite heavy occlusions between objects. To address this challenge, some approaches perform amodal reasoning in image space using foundation vision models to complete the occluded portions of objects \cite{Ardelean2025Gen3DSR}, while others defer this reasoning to the 3D reconstruction stage, either during geometry generation or by object retrieval \cite{Meng2025SceneGen,Wu2025Diorama, Gao2024DiffCAD,Yao2025CAST}. However, these approaches either operate primarily at the level of individual objects or do not fully leverage the repeated structure often present across multiple instances within a scene. In many real-world environments, objects commonly appear in sets, with multiple instances of the same object present in the scene. Observed from different viewpoints and with varying levels of occlusion, these instances provide complementary observations that can be leveraged to improve object reconstruction. 

To address these limitations, we propose FurnSet, which directly seeks to identify and exploit repeating object instances in 3D scenes. Our method first extracts per-object masks along with the corresponding segmented images using an off-the-shelf segmentation model \cite{Kirillov2023SAM}. The segmented outputs along with the scene image are then encoded with a pre-trained encoder to extract both scene-level and object-level visual features for model conditioning. To leverage repeating object instances, we introduce per-object CLS tokens \cite{Devlin2019BERT} and a similarity head to identify repeated object instances. Then our set-aware self-attention module effectively aggregates information from repeated instances into a single reconstruction, allowing each object instance to contribute where it is most spatially confident for the object. After generating 3D objects, our method estimates scene layout by first predicting a scene-level depth map \cite{Lin2025DepthAnything3,Yang2024DepthAnythingV2,Wang2025VGGT,Ke2024Marigold} and extracting per-object depth segments using their respective masks. These segments are then unprojected into point clouds and processed by our layout model. We parameterize the 3D object poses and optimize the spatial layout by minimizing the 3D and 2D projection losses between the generated object point clouds and the scene object point clouds, yielding the reconstructed 3D scene. In summary, our contributions are as follows:
\begin{itemize}
  \item We introduce a framework for single-view 3D reconstruction that effectively leverages both scene-level and object-level information to guide object geometry generation.
  \item We introduce CLS tokens and an object set-aware self-attention that identifies repeated object instances and jointly reconstructs their geometry by aggregating complementary observations across instances, improving reconstruction.
  \item We perform comprehensive evaluations validating the performance of our approach for single-view 3D reconstruction.
\end{itemize}

\section{Related Work}
\label{sec:relwork}
{\bf 3D Object Generation.} 3D object generation has seen significant advancement recently enabled by the release of extensive high-quality 3D object \cite{Chang2015ShapeNet,Deitke2023Objaverse,Deitke2023ObjaverseXL, Fu2021ThreeDFuture} and 3D scene \cite{Dai2017ScanNet,Yeshwanth2023ScanNetPP,Yu2025MetaScenes, Fu2021ThreeDFront} datasets. Several approaches leverage foundational image or video diffusion models \cite{Rombach2022LDM} to render multi-view images conditioned on single-view or sparse input images. These multi-view images are then used to reconstruct 3D scenes using representations such as Neural Radiance Fields (NeRFs) and 3D Gaussian Splatting (GS) \cite{Kerbl2023GaussianSplatting,Mildenhall2021NeRF}. Reconstruction is performed either by directly optimizing these representations from the generated views \cite{Gao2024CAT3D,Long2024Wonder3D,Liu2023SyncDreamer,Li2023Instant3D}, or by using feed-forward models to predict the parameters of the 3D representation \cite{Ardelean2025Gen3DSR}. Alternatively, some methods directly generate 3D representations, such as triplanes \cite{Peng2020CON,Wu2023Sin3DM,Wu2024Direct3D}, vector-set representations \cite{Zhang2023Shape2VecSet,Zhang2024CLAY,Hunyuan3DTeam2025Hunyuan3D21,Li2025TriPoSG}, or structured 3D latents \cite{Xiang2025StructuredLatents}. Triplane methods represent 3D features using three orthogonal feature planes that are queried to obtain features for a 3D location. Vector-set based methods sample query points from object point clouds with furthest point sampling. These query points then perform cross-attention with the object point cloud to obtain a representative latent code \cite{Zhang2023Shape2VecSet}. In contrast, structured 3D latent methods use sparse voxels with local latents defined at the active voxels intersecting the object’s surface \cite{Xiang2025StructuredLatents}. This yields better reconstruction quality, but requires a two-stage generation process to first generate the active voxels. Several approaches build on both the vector-set and structured latent based methods achieving strong results in amodal 3D object generation for occluded objects. Several approaches enhance the amodal reconstruction capabilities of structured latent models by fine-tuning them to be occlusion-aware \cite{Wu2025Amodal3R,Zhou2025AmodalGen3D, Chen2025SAM3D}. However, these methods remain limited by their focus on reconstructing a single object, which prevents them from leveraging amodal information available across multiple instances or contextual cues within the scene.

{\bf 3D Scene Generation.} 3D scene generation involves reconstructing the objects of a scene and determining their pose to reassemble a complete 3D scene. Retrieval-based methods typically use a combination of object detectors and feature extractors to identify the best matching CAD object to the one in the scene image \cite{Wu2025Diorama,Gao2024DiffCAD}. Although retrieved objects may be of high-quality and simulation ready, the reconstruction is constrained by the diversity and completeness of the available CAD database. Furthermore, in highly occluded and challenging scenes, the lack of visual information can result in mismatches between the retrieved model and the actual 3D object. Some generative methods complete the segmented object image using off-the-shelf pre-trained image models \cite{Ardelean2025Gen3DSR} like Stable Diffusion \cite{Rombach2022LDM}. The completed object image is then processed by an object reconstruction model to generate the 3D object. However, this handles the amodal reconstruction of each object independently, limiting the ability to leverage contextual information from the scene or shared structure across objects. Other methods instead perform amodal 3D reconstruction directly in 3D space \cite{Meng2025SceneGen,Yao2025CAST,Zhao2025DEPR}. However, they rely on scene context only implicitly and do not explicitly identify or exploit repeated object instances within the scene. This can lead to inaccurate reconstructions when objects are heavily occluded. To re-assemble the objects into the scene, several works leverage foundational depth models \cite{Lin2025DepthAnything3,Yang2024DepthAnythingV2,Wang2025VGGT} to extract 3D geometry from the scene image that can be passed in as features \cite{Meng2025SceneGen} or directly used to align objects into the scene \cite{Ardelean2025Gen3DSR,Wu2025Diorama}. Other methods instead jointly reason about object reconstruction and pose, but predicting the full scene layout for all objects simultaneously remains challenging in complex scenes. Some approaches instead condition full scene reconstruction directly on a single scene image \cite{Huang2025MIDI,Lin2025PartCrafter}. However, these methods often suffer from limited per-object reconstruction fidelity due to the use of a shared canonical representation for the entire scene.

\section{Methodology}
\label{sec:method}
Our proposed method reconstructs full 3D scenes from single image inputs, predicting the shape, texture, and position of scene objects. In Section 3.1, we introduce the problem formulation, in Section 3.2, we describe the set-aware structured voxel generation, in Section 3.3, we describe the generation of object shape and texture, in Section 3.4, we present the scene layout estimation strategy, and in Section 3.5, we detail the training of our method. 

\begin{figure}[H]
\centering
\includegraphics[width=\linewidth]{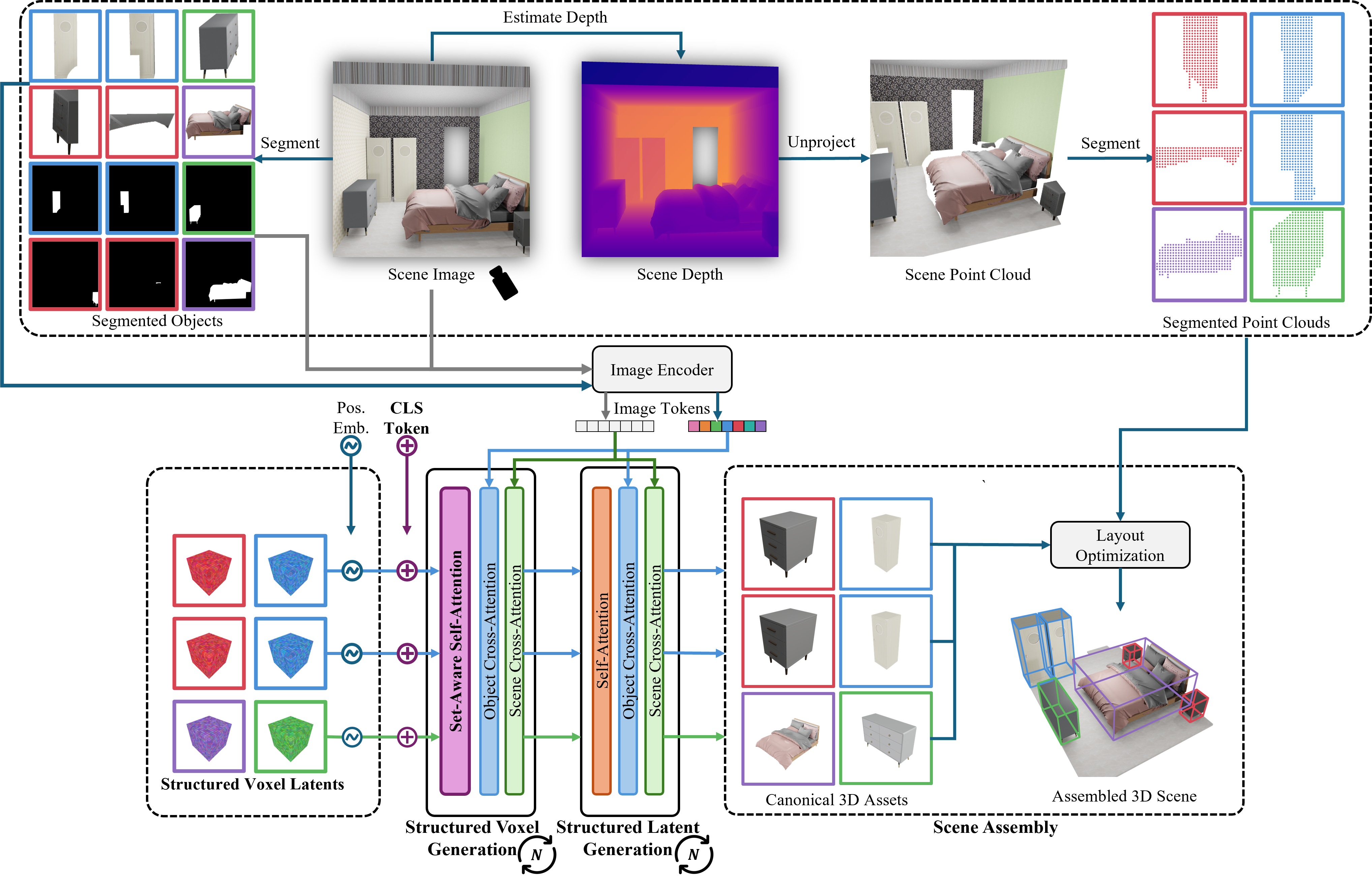}
\caption{{\bf Overview of the FurnSet Framework.} Our framework takes a single scene image and object segmentations as conditioning for object generation. During structured voxel generation, $CLS$ tokens are concatenated with object tokens to identify repeated instances, which are jointly reconstructed through \textit{set-aware self-attention}. The \textit{object cross-attention} and \textit{scene cross-attention} modules aggregate object-level and scene-level features extracted by the image encoder. The structured latent generation module synthesizes fine texture and appearance features. Finally, the generated objects are assembled into the scene using the predicted scene point cloud and layout optimization module.}
\label{fig:method}
\end{figure}

\subsection{Problem Formulation}
Given scene image $I_{scene}$ populated by $N$ objects composed of unique instances and sets of identical objects, our method outputs the corresponding structure, texture and position for each object $n_i$. We leverage an off-the-shelf segmentation model to obtain masked-out object images $\left\{I_i\right\}_{i=1}^N$, object masks $\left\{m_i\right\}_{i=1}^N$, and object point clouds $\left\{P_i\right\}_{i=1}^N$. Then the Structured Voxel Generation model ${\hat{G}}_S$ generates the structure $o_i$ of each object $n_i$ represented as sparse 3D voxels on an $R\times R\times R$ resolution 3D grid. To take advantage of the repetition found in indoor scenes, we assign a $CLS$ token to each object to identify recurring instances, thereby facilitating a joint reconstruction that ensures structural consistency. This enables better use of global scene context, ensuring identical objects mutually inform each other during generation. The texture and fine geometry of each object are then generated by the structured latent generation model ${\hat{G}}_L$. Given sparse 3D voxels $o_i$ for each object, the model ${\hat{G}}_L$ generates latents $z_i$, which are then decoded to 3D Gaussian Splat (GS) representations $O_i$ by decoder $D_{GS}$. 
Both ${\hat{G}}_S$ and ${\hat{G}}_L$ are conditioned on the scene image $I_{scene}$, the object masks $\left\{m_i\right\}_{i=1}^N$ and the object image $I_i$ for generating each 3D object. To place the generated 3D objects into the scene, a dedicated layout optimizer $F_{layout}$ is applied to deduce the pose $\phi_i$ of each instance: 
\begin{equation}
\phi_i = F_{\text{layout}}(O_i, P_i)
\label{eq:layout_embedding}
\end{equation}
The final scene reconstruction is given by the 3D reconstruction and pose of each object as $\left\{O_i,\phi_i\right\}_{i=1}^N$.

\subsection{Set-Aware Structured Voxel Generation}
As described in section 3.1, in the first stage we aim to generate the structure $o_i$ of each object $n_i$ represented as sparse 3D voxels on an $R\times R\times R$ resolution 3D grid. To generate the sparse structure latents for each object we leverage a rectified flow Diffusion Transformer (DiT \cite{Peebles2023DiT}). To identify sets of repeated object instances along with unique instances, we initialize a CLS token $c_i$ for each object. At every flow-matching step, this CLS token is concatenated with the object tokens and processed jointly through the DiT. Inspired by \cite{Dong2025CoPart,Ding2025FullPart,Meng2025SceneGen}, to capture intra-object information and scene-level information, we utilize both object-level self-attention and scene-level self-attention across all $N$ scene objects. The self-attention layer in the DiT blocks alternates between these two attention types, enabling both intra-object and global scene interactions. For object-level self-attention, attention for each object $n_i$ being generated is constrained to its respective object tokens $S_i$. 
\begin{equation}
z_{\text{self-obj}} = \text{Attention}(S_i, S_i, S_i)
\label{eq:self_obj_attention}
\end{equation}
For scene-level self-attention, attention is enabled between all object tokens simultaneously  $S=[S_1; S_2;\dots;S_N]$.
\begin{equation}
z_{\text{self-scene}} = \text{Attention}(S, S, S)
\label{eq:self_scene_attention}
\end{equation}
At each scene-level attention block, we first extract the $CLS$ token for each object from the current hidden states, yielding one summary token per object. Each $CLS$ token is then L2-normalized, and pairwise cosine similarity is computed between the $CLS$ tokens of all objects to produce a similarity matrix. This similarity matrix is used as a $CLS$-gated bias that modulates the strength of inter-object attention based on the learned object identity. Given $CLS$ tokens $\left\{c_i\right\}_{i=1}^N$, we compute pairwise similarity and use it to form an attention bias: 
\begin{equation}
s_{ij} = \sigma\!\left(\tau \cdot \operatorname{sim}_{\cos}(c_i, c_j)\right)
\label{eq:similarity_score}
\end{equation}

\begin{equation}
\alpha = \text{softmax}\left( \frac{QK^\top}{\sqrt{D}} + \log(s + \epsilon) \right)
\label{eq:attention_weights}
\end{equation}
where $c_i$ and $c_j$ denote the $CLS$ tokens of objects $i$ and $j$ and $\tau$ is a learnable temperature parameter. This formulation encourages repeated instances within a set to attend more strongly to one another than to other instances or object sets in the scene, promoting discovery of identical instances and sharing of geometry to mutually inform their reconstruction. The identity-aware biasing directs repeated instances within a set to jointly reconstruct using shared geometry across members. This yields richer and more complete geometry for repeated objects without conflating distinct ones.
Following the self-attention layers, to inject instance-specific visual evidence, we introduce an object-masked cross-attention layer between each object’s token set $S_i$ and image features extracted from the corresponding masked-out object image $I_i$. The object image $I_i$ is encoded with the same pre-trained image encoder DINOv2 to obtain per-object image features $F_i$. 
\begin{equation}
z_{\text{cross-obj}} = \text{Attention}(S_i, F_i, F_i)
\label{eq:cross_obj_attention}
\end{equation}
Following the object cross-attention layer, a scene-level cross-attention layer incorporates the scene image along with the object masks as conditioning where both are encoded with the pre-trained DINOv2 encoder to obtain image tokens $F_{scene}$.
\begin{equation}
z_{\text{cross-scene}} = \text{Attention}(S, F_{\text{scene}}, F_{\text{scene}})
\label{eq:cross_scene_attention}
\end{equation} 
The DiT generates the sparse structured voxels for each object through the flow matching process conditioned on the scene-level image, object masks, and masked-out object images.

\subsection{Structured Latent Generation}
The structured latent generation model ${\hat{G}}_L$ generates latent feature vectors $Z_{n_i}=\left\{z_i\right\}_{i=1}^N$ for the occupied voxels $V_{n_i}=\left\{v_i\right\}_{i=1}^N$ representing the sparse structure of object $n_i$. To generate the latents for the voxels $V_{n_i}$ of each object, the model ${\hat{G}}_L$ is conditioned on the segmented object mask and corresponding object image $I_i$. When a set of identical objects $S_k=\left\{n_i,\ n_j,\ ...\right\}$ is identified, instead of generating the latent features for the voxels of each instance, we share a single voxel structure $V_n$ for the set. Then at each cross-attention block in the structured latent generator, we compute the cross-attention for the mask $m_i$ and image $I_i$ of each object in the set separately. The outputs are then averaged before being passed to the subsequent transformer layer. The generated structured latents are then decoded with a pre-trained decoder $D_{GS}$ into textured 3D GS representations. The decoded object representations are then assembled to form the final 3D scene, with the single 3D GS object for a set being replicated and placed into the scene for each object in $S_k$.

\subsection{Layout Estimation}
To recover the scene layout for each object, we first utilize an off-the-shelf depth prediction model \cite{Lin2025DepthAnything3} to extract a per-pixel depth estimate $D_{scene}$ which is then unprojected into a point cloud $P_{scene}$. Point clouds from the surface of each object are extracted using masks $\left\{m_i\right\}_{i=1}^N$, yielding surface point clouds $\left\{P_i\right\}_{i=1}^N$. The pose of each object in the scene is parameterized as a set of spatial transformation parameters $\phi_i$ by translation (3 parameters), rotation (1 parameter) around the vertical axis, and scale (1 parameter). The parameters of each object are optimized via gradient descent to achieve alignment between the 3D objects and the image scene. In addition to the optimization objective of minimizing the 3D Chamfer Distance (CD) loss between the object point cloud $P_{n_i}$ and the observed surface point cloud $P_i$, a 2D projection constraint is also used. The 2D CD is computed between $P_{n_i}$ and $P_i$ projected to the image plane with known intrinsics. The full scene scale, position, and orientation are then recovered for each object $n_i$  with the optimized parameters $\phi_i$. The overall optimization function is defined as: \begin{equation}
\mathcal{L}_{\text{layout}} = \lambda_1 \cdot \mathcal{L}_{{CD}}^{3D}(P_{n_i}, P_i) + \lambda_2 \cdot \mathcal{L}_{{CD}}^{2D}({Proj}(P_{n_i}), {Proj}(P_i))
\label{eq:layout_loss}
\end{equation}
where $\lambda_1$ and $\lambda_2$ are weights for the 3D and 2D CD terms respectively, and $Proj(\cdot)$ denotes the projection to the image plane.

\subsection{Training}
We train our structured voxel DiT using the Conditional Flow Matching (CFM) loss \cite{Lipman2023FlowMatching}, where the CFM objective trains a neural network $v_\theta\left(x,t\right)$ to predict the velocity field that moves a noisy sample $x$ towards a target data distribution: \begin{equation}
\mathcal{L}_{\text{CFM}}(\theta) = \mathbb{E}_{t, x_0, \epsilon} \left\| v_\theta(x, t) - (\epsilon - x_0) \right\|_2^2
\label{eq:cfm_loss}
\end{equation}
where $\epsilon$ is the noise and $x_0$ is the clean data sample. To train the $CLS$ tokens to produce meaningful pairwise similarities for the identity-aware attention bias, we supervise them with a pairwise similarity loss. At the output of the final transformer block, each $CLS$ token $c_i$ is projected through an MLP head and L2-normalized. For every pair of objects $(i,j)$ in the scene, we compute their cosine similarity and supervise it with a binary cross-entropy loss against the ground-truth identity label $y_{ij} \in \{0,1\}$, where $y_{ij}=1$ if objects $i$ and $j$ belong to the same repeated-instance set: 

\begin{equation}
\mathcal{L}_{\text{sim}} = -\frac{1}{|\mathcal{R}|} \sum_{(i,j) \in \mathcal{R}} \left[ y_{ij} \log(\hat{s}_{ij}) +
(1 - y_{ij}) \log(1 - \hat{s}_{ij}) \right]
\label{eq:sim_loss}
\end{equation}  
where $\hat{s}_{ij} = \sigma(c_i^\top c_j / \tau_c)$ is the predicted similarity, $\sigma$ denotes the sigmoid function, $\tau_c$ is a temperature hyperparameter, and $\mathcal{R}$ is the set of all object pairs in the scene. This formulation directly supervises every object pair, including pairs of singleton objects, ensuring that the CLS similarity signal remains reliable across all objects in the scene. The total training loss combines flow matching and similarity supervision:
\begin{equation}
\mathcal{L} = \mathcal{L}_{\text{CFM}} + \lambda_s \mathcal{L}_{\text{sim}}
\label{eq:total_loss}
\end{equation}

\section{Experiments}
\label{sec:exp}
\subsection{Experiment Settings}
{\bf Datasets and Baselines.} We conduct experiments on the 3D-Front and 3D-Future datasets \cite{Fu2021ThreeDFuture,Fu2021ThreeDFront}. The 3D-Future dataset contains 9,992 objects and 20,240 scenes. To enhance our dataset with more scenes that contain repeated object instances, we leverage the 3D-Front dataset as well. The 3D-Front dataset leverages the object meshes from 3D-Future and constructs further realistic scene layouts. We leverage the scenes from the 3D-Future training dataset along with 12,000 images collected from the 3D-Front dataset. For validation, we use 200 scene images combined from both test datasets. Qualitative and quantitative comparisons are performed between our method and MIDI \cite{Huang2025MIDI}, SceneGen \cite{Meng2025SceneGen}, and SAM-3D \cite{Chen2025SAM3D}.

{\bf Metrics.} To evaluate the geometric structure of the generated scenes and objects, we first reconstruct point clouds from the synthesized asset surfaces and align them with the ground truth scenes and objects using FilterReg \cite{Gao2019FilterReg}. We then compute scene-level Chamfer Distance (CD-S) and F-Score (F-Score-S), object-level Chamfer Distance (CD-O) and F-Score (F-Score-O). To demonstrate the impact of our set-aware generation, we additionally report CD-O and F-Score-O restricted to repeated object instances within a scene.

{\bf Implementation Details.} Our model is trained on 8 A100 GPUs with batch size 8. We implement classifier-free guidance (CFG \cite{Ho2022CFG}) with a drop rate of 0.1 and the AdamW \cite{Loshchilov2017AdamW} optimizer with a learning rate of 1e-4. At inference, we use 25 sampling steps with the classifier-free guidance (CFG) scale set to w=5.0.

\subsection{Quantitative Results}
As seen in Table 1, we compare our method against MIDI, SceneGen, and SAM-3D, where our approach achieves the best overall performance. Our method consistently outperforms existing approaches across a range of challenging indoor scenes, demonstrating stronger robustness to occlusion and improved reconstruction quality. Notably, the performance gains are most pronounced on scenes containing repeated object instances, where our method effectively identifies and uses shared structure across instances to produce more accurate reconstructions. These results highlight the advantage of explicitly modeling object repetition, enabling our framework to better leverage complementary observations compared to methods that treat objects independently.
\begin{table}[H]    
\centering
\scriptsize
\setlength{\tabcolsep}{3pt}
\renewcommand{\arraystretch}{1.1}
\resizebox{\linewidth}{!}{%
\begin{tabular}{l|cccc|cc}
\toprule
\multirow{2}{*}{\textbf{Method}}
& \multicolumn{4}{c|}{\textbf{3D-Future \& 3D-Front}}
& \multicolumn{2}{c}{\textbf{Repeated Instances}} \\
& \textbf{CD-S$\downarrow$} & \textbf{CD-O$\downarrow$} & \textbf{F-Score-S$\uparrow$} & \textbf{F-Score-O$\uparrow$}
& \textbf{CD-O$\downarrow$} & \textbf{F-Score-O$\uparrow$} \\
\midrule
MIDI \cite{Huang2025MIDI}
& 0.0628 & 0.0647 & 63.08 & 39.28
& 0.0634 & 39.92 \\

SceneGen \cite{Meng2025SceneGen}
& 0.0564 & 0.0483 & 66.89 & 62.09
& 0.0458 & 63.21 \\

SAM-3D \cite{Chen2025SAM3D}
& 0.0584 & 0.0437 & 64.71 & 65.46
& 0.0443 & 64.27 \\

FurnSet (Ours)
& \textbf{0.0451} & \textbf{0.0336} & \textbf{75.13} & \textbf{74.14}
& \textbf{0.0281} & \textbf{77.72} \\
\bottomrule
\end{tabular}%
}
\caption{Quantitative comparison on 3D-Future and 3D-Front datasets. Additional results exclusively for repeated object instances are also reported.}
\label{tab:quant_table}
\end{table}

\subsection{Qualitative Results}
As seen in Figures 2 and 3, our model effectively identifies identical object sets and leverages the scene information to jointly reconstruct the set. This yields significant improvements in geometric fidelity, allowing the visual conditioning from each object to optimally refine a shared structural template. Specifically, aggregating these features across the set resolves ambiguities caused by occlusions or sparse inputs in any single instance. Furthermore, this is achieved while maintaining strong reconstruction of independent, non-set object instances. 

\begin{figure}[H]
\centering
\includegraphics[width=0.8\linewidth]{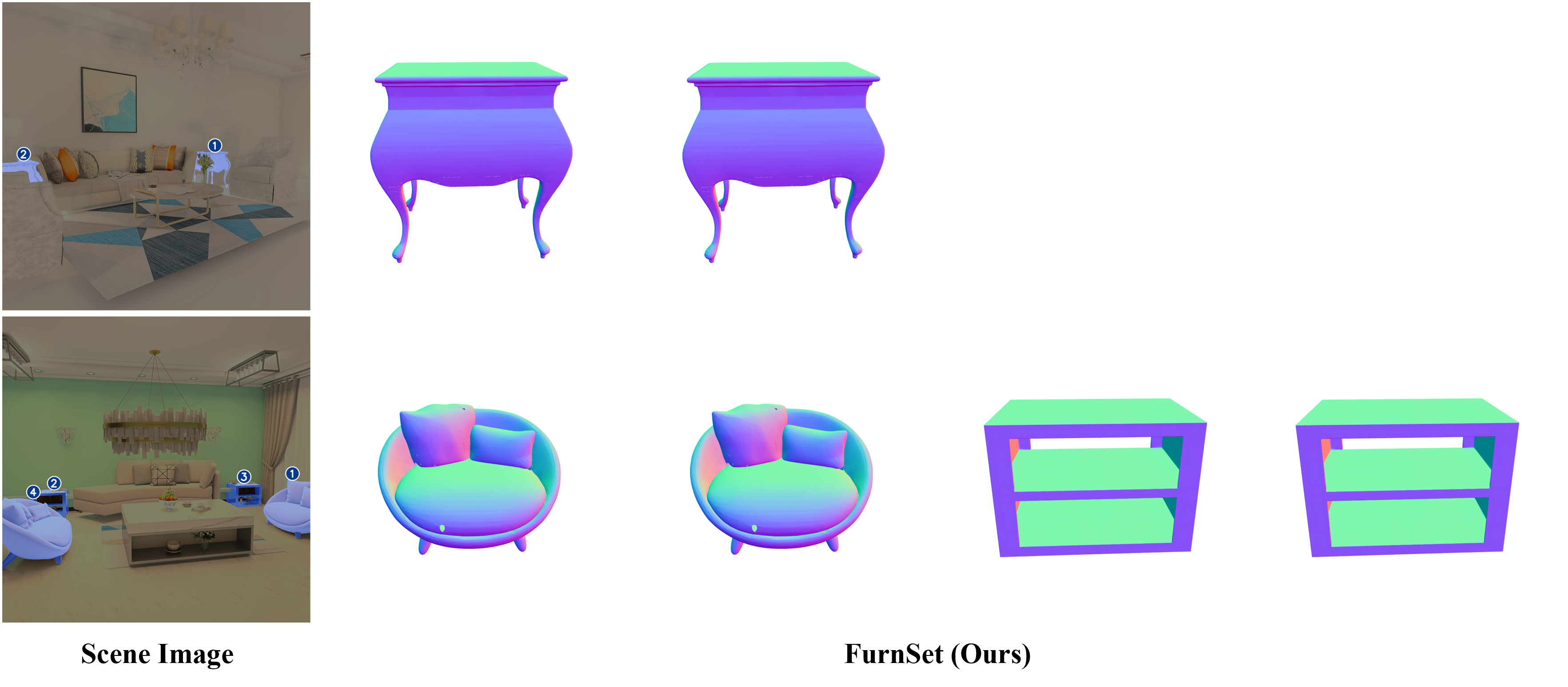}
\caption{Repeated object set reconstruction under occlusion.}
\label{fig:objects}
\end{figure}

\begin{figure}[H]
\centering
\includegraphics[width=0.8\linewidth]{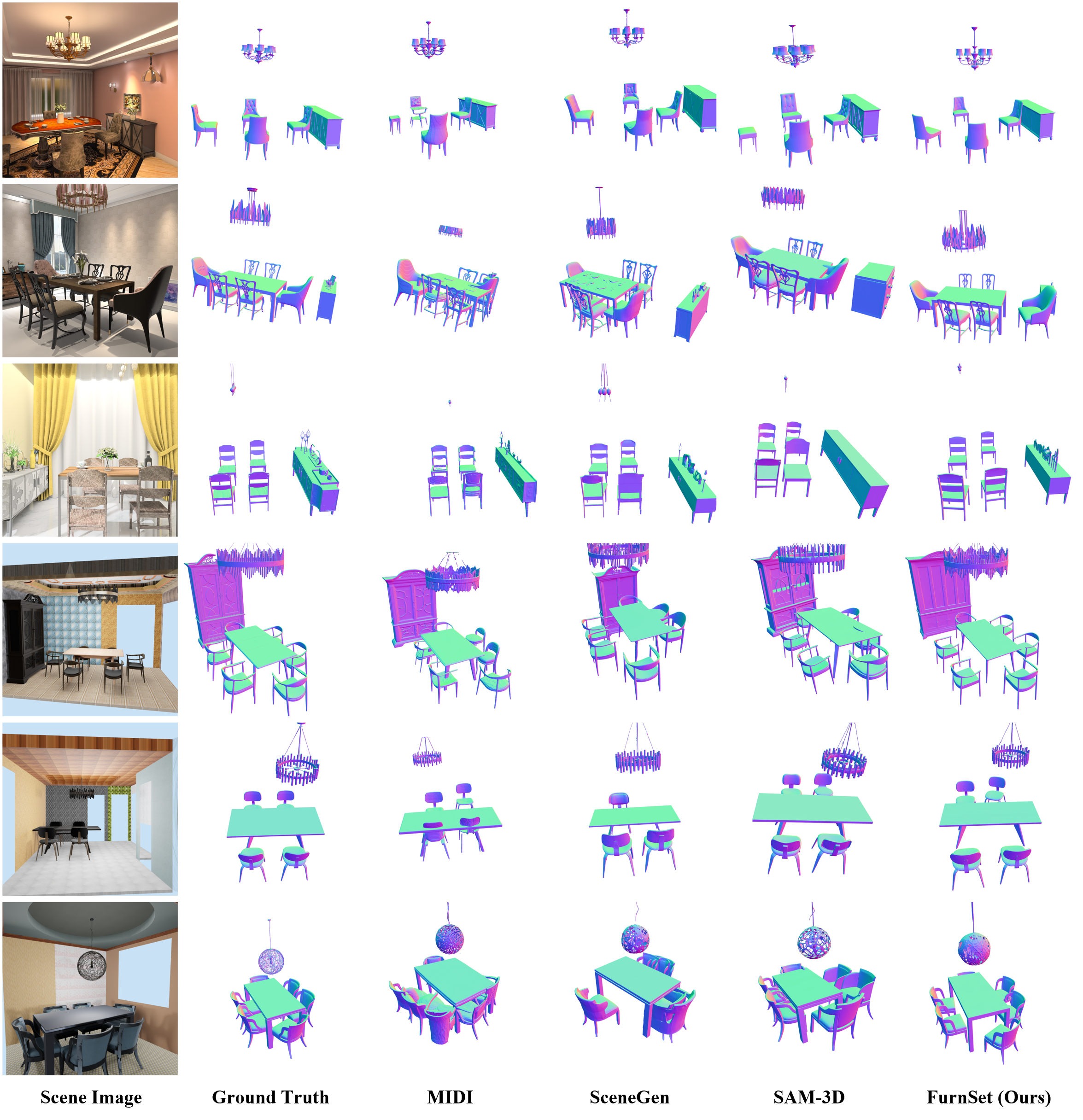}
\caption{Qualitative comparison with scene generation methods. The top three scenes are from the 3D-Future dataset \cite{Fu2021ThreeDFuture} and the bottom three scenes are from the 3D-Front dataset \cite{Fu2021ThreeDFront}.}
\label{fig:comparison}
\end{figure}

\subsection{Ablation Study}
\textbf{Global Self-Attention.} We ablate the impact of global scene-level attention by restricting the model to object-level self-attention only, where each object's tokens attend exclusively to themselves with no cross-object interaction. Without global attention, each object is generated largely independently, relying solely on its own conditioning image and the shared scene image and object masks. The model cannot leverage global scene structure or spatial context from other objects to inform its reconstruction, limiting its ability to resolve ambiguities in regions where a single object's conditioning provides insufficient information.

\textbf{Set-Aware Global Self-Attention.} We further ablate the set-aware self-attention by replacing it with standard unbiased scene-level self-attention and removing the associated CLS tokens used for set identification. This removes the identity-driven attention modulation that suppresses attention between non-identical objects. Without this bias, attention between repeated instances is no longer preferentially emphasized relative to unrelated objects, reducing the model’s ability to effectively direct shared geometric information within a set. As a result, the model is less able to jointly leverage conditioning signals from repeated objects. Repeated instances therefore benefit less from joint reconstruction and exhibit greater geometric inconsistencies across set members, particularly under heavy occlusion.

\begin{table}[H]
\centering
\small
\setlength{\tabcolsep}{6pt}
\renewcommand{\arraystretch}{1.1}
\begin{tabular}{l|cccc|cc}
\toprule
\multirow{2}{*}{\textbf{Method}}
& \multicolumn{4}{c|}{\textbf{3D-Future \& 3D-Front}}
& \multicolumn{2}{c}{\textbf{Repeated Instances}} \\
& \textbf{CD-S$\downarrow$} & \textbf{CD-O$\downarrow$} & \textbf{F-Score-S$\uparrow$} & \textbf{F-Score-O$\uparrow$}
& \textbf{CD-O$\downarrow$} & \textbf{F-Score-O$\uparrow$} \\
\midrule
w/o GSA      & 0.0596 & 0.0540 & 64.53 & 57.38 & 0.0553 & 56.17 \\
w/o Set-Aware GSA  & 0.0522 & 0.0501 & 71.27 & 61.17 & 0.0490 & 62.08 \\
FurnSet            & \textbf{0.0451} & \textbf{0.0336} & \textbf{75.13} & \textbf{74.14} & \textbf{0.0281} & \textbf{77.72} \\
\bottomrule
\end{tabular}
\caption{Ablation study results.}
\label{tab:ablation}
\end{table}

\section{Conclusion}
\label{sec:conc}
In this paper we presented FurnSet, a framework for single-view 3D scene reconstruction that explicitly leverages repeated object instances to improve reconstruction under occlusion. Given a single scene image, our method jointly reconstructs object geometry and estimates spatial layout using both scene-level and object-level conditioning. We introduce per-object CLS tokens and a set-aware self-attention mechanism that identifies repeated instances and aggregates complementary observations for joint reconstruction. Experiments on 3D-Future and 3D-Front demonstrate strong performance, with notable improvements on repeated object instances, highlighting the effectiveness of exploiting repetition for robust 3D scene reconstruction.

\clearpage
\bibliography{egbib}

\end{document}